\documentclass[]{informs3} 
\OneAndAHalfSpacedXII 

\usepackage{natbib}
\usepackage[ruled]{algorithm}
\usepackage[noend]{algorithmic}
 \bibpunct[, ]{(}{)}{,}{a}{}{,}%
 \def\bibfont{\small}%
\allowdisplaybreaks 

\newcommand{\CUT}[1]{{}}

\renewcommand{\paragraph}[1]{\smallskip \noindent {\textbf{#1}}}

\input epsf

\newtheorem{theorem}{Theorem}

\newtheorem{assumption}{Assumption}

\newtheorem{corollary}[theorem]{Corollary}
\newtheorem{definition}{Definition}
\newtheorem{lemma}{Lemma}

\begin{document}




 \TITLE{Constrained Stochastic Submodular Maximization with State-Dependent Costs}

\ARTICLEAUTHORS{%
\AUTHOR{Shaojie Tang}
\AFF{University of Texas at Dallas}
} 

\ABSTRACT{%
In this paper, we study the constrained stochastic submodular maximization problem with state-dependent costs. The input of our problem is a set of items whose states (i.e., the marginal contribution and the cost of an item) are drawn from a known probability distribution. The only way to know the realized state of an item is to select that item. We consider two constraints, i.e., \emph{inner} and \emph{outer} constraints. Recall that each item has a state-dependent cost, and the inner constraint states that the total \emph{realized} cost of all selected items must not exceed a give budget.  Thus, inner constraint is state-dependent. The outer constraint, one the other hand, is state-independent. It can be represented as a downward-closed family of  sets of selected items regardless of their states. Our objective is  to maximize the objective function  subject to both inner and outer constraints. Under the assumption that larger cost indicates larger ``utility'', we present a constant approximate solution to this problem.}


\maketitle

%

\section{Introduction}
\label{sec:intro}
In this paper, we study a novel constrained stochastic submodular maximization problem. We follow the framework developed in \citep{fukunaga2019stochastic} and introduce the state-dependent item costs into the classic stochastic submodular maximization problem. The input of our problem is  a set of items, each item has a random state which is drawn from a known probability distribution. The marginal contribution and the cost of an item are dependent on its actual state.  The utility function is a mapping from sets of items and their states to a real number. We must select an item before observing its actual state. Our objective is to sequentially select a group of items to maximize the objective function.  We must obey two constraints, namely, inner and outer constraints, through the selection process. The inner constraint requires that the total \emph{realized} cost of all selected items must not exceed a given budget  $B$. Thus, the inner constraint is state-dependent. The outer constraint is represented as a downward-closed family of  sets of selected items regardless of their states. Thus, the outer constraint is state-independent. Under the assumption that the cost of an item is larger if it is in a ``better'' state, we present a constant approximate solution.

Our model is general enough to capture many real-world applications. Here we discuss three examples.

\textbf{Adaptive coupon allocation.} 
The objective of this problem \cite{tang2018stochastic} is to distribute coupons to a group of up to $k$ seed users such that those who redeem the coupon can generate the largest cascade of influence. The state, which is stochastic, of a user is her decision on whether or not to redeem the coupon.  
In this case, it is uncertain in advance how many users redeem the coupon and help to promote the product. Our framework can capture this scenario by treating $k$ as a outer constraint and treating the budget on the total value of redeemed coupons  as a inner constraint.

\textbf{Recruiting crowd workers.} Crowdsourcing \cite{carvalho2016many} is an effective way of obtaining information from a large group of workers. A typical crowdsourcing process can be described as follows: the task-owner sequentially hires up to $k$ workers to work on a set of similar tasks. Each worker reports her results to the task-owner after completing her task. The state of a worker is the quality of the results returned by her. It is clearly reasonable  to assume that the actual amount of reward paid to a particular worker depends on her state, which can only be observed after she delivers the task. By treating $k$ as a outer constraint and treating the budget on the total value of total payments as a inner constraint, our objective is to maximize the overall quality of the completed tasks subject to outer and inner constraints.

\textbf{Recommendation.} In the context of product or news recommendation, our objective is to recommend a group of items to a customer in order to maximize some utility function which often satisfies the diminishing marginal return property. The performance  of a recommended item depends on many random factors such as the customer's preferences. For example, after receiving a recommended article, the customer decides to skip or read it in a probabilistic manner and she must spend her own resource such as time and money on reading a recommended article. Our framework can capture this scenario by treating the customer's decision on a particular article as the state of that article. Hence, the performance, as well as the cost, of a recommended article is determined by its state.

\emph{Related works.} Stochastic submodular maximization has been extensively studied in the literature \citep{golovin2011adaptive,chen2013near,fujii2016budgeted}. While most of existing works assume that the cost of each item is deterministic and pre-known, we consider state-dependent item costs. Our problem reduces to the stochastic knapsack problem \citep{gupta2011approximation} when considering linear objective function.  Recently, \cite{fukunaga2019stochastic,tang2021aaim} extended the previous study to the stochastic submodular maximization problem, however, their model does not incorporate outer constraints. Hence, our study can be considered as an extension of \citep{fukunaga2019stochastic}.  Our work is also closely related to submodular probing problem \citep{adamczyk2016submodular} where they assume each item  has binary states, our model allows each item to have multiple states.


\section{Preliminaries and Problem Formulation}
\label{sec:behavior}
 \paragraph{Lattice-submodular functions} Let $I=\{1,2,\cdots, n\}$ be a set of items and $[B]=\{1,2,\cdots, B\}$ be a set of states. We further define $[0;B]=\{0, 1,2,\cdots, B\}$.  Given two vectors $u,v \in [0;B]^I$,  $u\leq v$ means that $u(i)\leq v(i)$ for all $i\in I$. Define $(u\vee v)(i)=\max\{u(i),v(i)\}$ and $(u\wedge v)(i)=\min\{u(i),v(i)\}$. For $i\in I$, define $\mathbf{1}_i$ as the vector that has a $1$ in the $i$-th coordinate and $0$ in all other coordinates.  A function $f: [0;B]^I\rightarrow \mathbb{R}_{+}$ is called \emph{monotone} if $f(u)\leq f(v)$ holds for any $u, v\in [0;B]^I$ such that $u\leq v$, and $f$ is called \emph{lattice submodular} if $f(u\vee s\mathbf{1}_i)-f(u) \geq f(v\vee s\mathbf{1}_i)-f(v)$ holding for any $u,v\in [0;B]^I$, $s\in [0;B]$, $i\in I$.

 \paragraph{Items and States } We use a vector $\phi\in [B]^I$ to denote a \emph{realization} where for each item $i\in I$, $\phi(i)\in [B]$ denotes the state of  $i$ under realization $\phi$. We assume that there is a known prior probability distribution $p_i$ over realizations for each item $i$, i.e., $p_i=\{\Pr[\phi(i) = s]: s\in [B]\}$.  The states of all items are decided independently at random, i.e., $\phi$ is drawn randomly from the product distribution $p=\prod_{i\in I}p_i$. For each item $i\in I$ and state $s\in [B]$, let $c_i(s)$ denote the cost of  $i$ when its state is $s$. We made the following assumption.
\begin{assumption} For all $i\in I$ and $s, s'\in [B]$ such that $s \geq s'$, we have $c_i(s)\geq c_i(s')$, i.e., the cost of an item is larger if it is in a ``better'' state.
\end{assumption}


\paragraph{Adaptive Policy and Problem Formulation} 
Formally, a policy $\pi$ is a function that specifies which item to select next based on the observations made so far.  Consider any $S\subseteq I$ and any realization $\phi$, we use $\phi_S$ to denote a vector in $[0;B]^I$ such that $\phi_S(i)=\phi(i)$ if $i\in S$, and $\phi_S(i)=0$ otherwise. The utility of $S$ conditioned on $\phi$ is $f(\phi_S)$ where $f: [0;B]^I \rightarrow \mathbb{R}_+$ is a monotone and lattice-submodular function. Consider an arbitrary policy $\pi$, for each $\phi$, let $I(\pi, \phi)$ denote the set of items selected by $\pi$ conditional on $\phi$ \footnote{For simplicity, we only consider deterministic policy. However, all results can be easily extended to random policies.}. Let $\Phi$ denote a random realization, the expected utility of $\pi$ is written as
\begin{eqnarray}
f_{avg}(\pi)=\mathbb{E}_{\Phi\sim p} [f(\Phi_{I(\pi, \Phi)}))]
\end{eqnarray}

Moreover, for any subset of items $S\subseteq I$, define $\overline{f}(S)= \mathbb{E}_{\Phi\sim p}[f(\Phi_S)]$ as the expected utility of $S$ with respect to the distribution $p$.

\begin{definition}
We say a policy $\pi$ is \emph{feasible} if it satisfies both outer and inner constraints:
\begin{enumerate}
\item(Inner Constraint)  For all $\phi$, we have $\sum_{i\in I(\pi, \phi)}c_i(\phi(i))\leq C$.
\item(Outer Constraint) For all $\phi$, we have $I(\pi, \phi)\in \mathcal{I}^{out}$, where $\mathcal{I}^{out}$ is a downward-closed family of sets of items.
\end{enumerate}
\end{definition}
Our goal is to identify the best feasible policy that maximizes its expected utility.
\[\max_{\pi} f_{avg}(\pi) \mbox{ subject to } \forall \phi: \sum_{i\in I(\pi, \phi)}c_i(\phi(i))\leq C; I(\pi, \phi)\in \mathcal{I}^{out}.\]

Following the framework  developed in \cite{chekuri2014submodular,fukunaga2019stochastic}, our algorithm is composed of two phases, a continuous optimization phase and a rounding phase. We first solve a continuous optimization problem and obtain a fractional solution. In the rounding phase, we convert the continuous solution to a feasible adaptive policy that obeys inner and outer constraints. We first explain the continuous optimization phase.

\section{Continuous Optimization Phase}
We present our solution based on the concept of ``time''. We assume that each item $i$ is associated with a random processing time $c_i(\phi(i))$. Hence, if an item $i$ is selected at time $t$, we must wait until $t+c_i(\phi(i))$ to select the next item. We treat the budget $C$ as the time limit. We can not select an item $i$ at slot $t$ if the processing of $i$ may finish before a time limit $C$, i.e., $i$ can not be selected at time $t$ if  there exists some state $s\in [B]$ such that $t+c_i(s)> C$. We define a variable $x(i,t)$ for each item $i$ and slot $t$, and it indicates whether an item $i$ is selected at $t$. Let $\overline{x}\in \mathbb{R}_+^I$ denote a vector defined by $\overline{x}(i)=\sum_{t\in[C-c_i(B)]}x(i,t)$.  We next introduce the multilinear extension $F(\overline{x})$ of $\overline{f}$. \[F(\overline{x})=\sum_{U\subseteq I} \prod_{i\in U} \overline{x}(i)\prod_{i \notin U} (1-\overline{x}(i))\overline{f}(U)\]

Let $P_{\mathcal{I}^{out}} \subseteq [0,1]^I$  denote a polytope that is a  relaxation for $\mathcal{I}^{out}\subseteq 2^I$, i.e.,  $P_{\mathcal{I}^{out}} = \mathrm{conv}\{\mathbf{1}_{S}\mid S\in \mathcal{I}^{out}\}$. For any $d\in [0,1]$, let $d \cdot P_{\mathcal{I}^{out}}=\{d  \cdot \overline{x}\mid \overline{x}\in P_{\mathcal{I}^{out}}\}$. Now we are ready to introduce the continuous optimization problem as follows:
\begin{center}
\framebox[0.6\textwidth][c]{
\enspace
\begin{minipage}[t]{0.6\textwidth}
\small
\textbf{P1:}
\emph{Maximize $F(\overline{x})$ }\\
\textbf{subject to:}
\begin{equation*}
\begin{cases}
\forall i\in I: \overline{x}(i)=\sum_{t\in[C-c_i(B)]}x(i,t)\\
\forall i\in I: \overline{x}(i)\leq 1\\
\overline{x}\in P_{\mathcal{I}^{out}} \\
\forall t\in [C]: \sum_{i\in I}\mathbb{E}[\min\{c_i(\phi(i)), t\}]\sum_{t'\in[t]}x(i,t')\leq 2t \quad \mbox{(C1)}
\end{cases}
\end{equation*}
\end{minipage}
}
\end{center}
\vspace{0.1in}

In constraint (C1), the expectation $\mathbb{E}[\min\{c_i(\phi(i)), t\}]$ is taken with respect to  $p$, i.e.,  $\mathbb{E}[\min\{c_i(\phi(i)), t\}]=\sum_{s=1}^B p_i(s)\min\{c_i(s), t\}$. Note that the formulation of \textbf{P1} involves $\Omega(I\times C)$ variables, which makes our algorithm pseudo polynomial. However,   we can apply the
technique used in \cite{gupta2011approximation} to convert the algorithm into a polynomial-time algorithm  at the expense weakening the approximation ratio by a constant factor. We can adopt the stochastic continuous greedy algorithm developed in \cite{asadpour2016maximizing} to solve \textbf{P1}. Their algorithm involves two controlling parameters: stopping time $l \in[0,1]$ and step size $\delta$. Their original analysis can be easily extended to show that for a stopping time $l\in[0,1]$, the algorithm outputs a solution $x$ such that $\overline{x}\in l \cdot P_{\mathcal{I}^{out}}$,  $\forall t\in [C]: \sum_{i\in I}\mathbb{E}[\min\{c_i(\phi(i)), t\}]\sum_{t'\in[t]}x(i,t')\leq  l\cdot 2t$, and  $F(\overline{x})\geq (1-e^{-l}-O(n^3\delta))f_{avg}(\pi^{opt})$ where  $\pi^{opt}$ denotes the optimal policy of our original problem. The following lemma  follows immediately from the above observation.

\begin{lemma}
\label{lem:kao}
Let $\pi^{opt}$ denote the optimal policy of our original problem. If we apply the stochastic continuous greedy algorithm
with stopping time $l\in[0,1]$ and step size $\delta= o(n^{-3})$ to solve \textbf{P1}, then the algorithm outputs a solution
$x$ such that $\overline{x}\in l \cdot P_{\mathcal{I}^{out}}$,   $\forall t\in [C]: \sum_{i\in I}\mathbb{E}[\min\{c_i(\phi(i)), t\}]\sum_{t'\in[t]}x(i,t')\leq  l\cdot 2t$ and  $F(\overline{x})\geq (1-e^{-l}-o(1))f_{avg}(\pi^{opt})$.
\end{lemma}



\section{Rounding Phase}
In this section, we introduce an effective rounding approach that converts the continuous solution to an adaptive policy.
Before explaining the rounding phase, we first introduce two important concepts: \emph{$(\beta, \gamma)$-balanced contention resolution scheme} \cite{chekuri2014submodular} and \emph{$\alpha$-contention resolution scheme} \cite{fukunaga2019stochastic}.
\subsection{Contention Resolution Scheme}$(\beta, \gamma)$-balanced contention resolution scheme ($(\beta, \gamma)$-balanced CRS)  is a general framework designed for maximizing set-submodular functions. In \cite{fukunaga2019stochastic}, the authors extend this concept to the lattice-submodular functions by introducing $\alpha$-contention resolution scheme ($\alpha$-CRS).

 We first introduce the concept of $(\beta, \gamma)$-balanced CRS.
 \begin{definition}[$(\beta, \gamma)$-balanced CRS]
Given a vector $\overline{z}\in \beta \cdot P_{\mathcal{I}^{out}}$, let $R$ denote a random set of $I$ obtained by including each item $i\in I$ independently with probability $\overline{z}(i)$. A $(\beta, \gamma)$-balanced CRS with regards to $\overline{z}$ is a mapping $\chi : 2^I \rightarrow \mathcal{I}^{out}$ such that $\Pr[i\in \chi(R)| i\in R] \geq \gamma$, where the probability considers two sources of randomness: one is the randomness in choosing $R$, and the other source is the randomness in the execution of $\chi$. A $(\beta, \gamma)$-balanced CRS  is said to be monotone if for any two sets $R, R'$ such that  $R \subseteq R'$, the following inequality holds: $\Pr[i\in \chi(R)| i\in R] \geq \Pr[i\in \chi(R')| i\in R']$.
\end{definition}

We next introduce the concept of $\alpha$-contention resolution scheme. Consider a probability distribution $q: I\times [0;B]\rightarrow [0,1]$. Let $v\in[0;B]^I$ denote a random vector such
that, for each $i\in I$, the value of $v(i)$ is
set to  $j \in [0;B]$ independently with probability $q(i, j)$.


 \begin{definition}[$\alpha$-CRS] Let
 $\mathcal{F}\subseteq [0;B]^I$ be a downward-closed subset of $[0;B]^I$, that is, $u\leq v \in \mathcal{F}$ implies $u\in \mathcal{F}$, and let $\alpha\in [0, 1]$. An $\alpha$-contention resolution scheme ($\alpha$-CRS) with regards to $q$ is a mapping $\psi : [0;B]^I\rightarrow \mathcal{F}$ that  satisfies the following two conditions:
\begin{itemize}
\item For each  $i\in I$, $\psi(v)(i) \in\{0, v(i)\}$;
\item For each $i\in I$ and each $j\in [B]$, we have $\Pr[\psi(v)(i) = j| v(i)=j] \geq \alpha$, where the probability considers two sources of randomness: one is the randomness in choosing $v$, and the other source is the randomness in the execution of $\psi$.
\end{itemize}

An $\alpha$-CRS $\psi$ is said to be monotone if, for each $u, v \in [0;B]^I$
such that $u(i) = v(i)$ and $u \leq v$, we have $\Pr[\psi(u)(i) = u(i)] \geq \Pr[\psi(v)(i) = v(i)] $, where the probability here considers only the randomness in the execution of $\psi$.
\end{definition}

%

In the context of maximizing set-submodular functions,  Lemma 1.6 in \cite{chekuri2014submodular} states that one can combine contention resolution schemes for different constraints. We next follow a similar proof of theirs to show that this result also holds for lattice-submodular functions.
\begin{lemma}
\label{lem:new4}Let $\mathcal{F}=\bigcap_{t=1}^k\mathcal{F}^t$ denote the intersection of several different subsets of $[0;B]^I$ where for each $t\in [k]$, $\mathcal{F}^t\subseteq [0;B]^I$ is a downward-closed subset of $[0;B]^I$. Suppose each $\mathcal{F}^t$ has a monotone $\alpha_t$-CRS with regards to $q$. Then $\mathcal{F}$ has  a monotone $\prod_{t=1}^k\alpha_t$-CRS with regards to $q$.
\end{lemma}

\emph{Proof:} We assume $k = 2$ for simplicity; the general statement can be proved by induction. Given a vector $v\in [0;B]^I$, for each $t\in\{1, 2\}$, assume that we can apply a monotone $\alpha_t$-CRS  $\psi^t$ separately to obtain $\psi^t(v)$. Then we define a mapping $\psi : [0;B]^I\rightarrow \mathcal{F}$ such that
\begin{equation*}\mbox{for each } i\in I: \psi(v)(i)=
\begin{cases}
v(i)\quad &\mbox{if for all } t\in \{1,2\}, \psi^t(v)(i)=v(i)\\
0 \quad &\mbox{otherwise }
\end{cases}
\end{equation*}
We next show that $\psi$ is a monotone $\alpha_1\alpha_2$-CRS with regards to $q$.
Conditioned on $v$, the value of $\psi^1(v)$,  $\psi^2(v)$ are independent, which means that
\begin{eqnarray}
\Pr[\psi(v)(i) = v(i)] &=& \Pr[\psi^1(v)(i) = v(i) \&\psi^2(v)(i) = v(i)] \\
&=& \Pr[\psi^1(v)(i) = v(i)] \Pr[\psi^2(v)(i) = v(i)]
\end{eqnarray}

Taking an expectation over $v$ conditioned on $v(i)=j$, we get
\begin{eqnarray}\Pr[\psi(v)(i) = j| v(i)=j] &=& \mathbb{E}_{v\sim q}[ \Pr[\psi(v)(i) = j]| v(i)=j] \\
&=& \mathbb{E}_{v\sim q}[ \Pr[\psi^1(v)(i) = j\&\psi^2(v)(i) = j]| v(i)=j] \\
&=& \mathbb{E}_{v\sim q}[\Pr[\psi^1(v)(i) = j] \Pr[\psi^2(v)(i) = j]| v(i)=j] \label{eq:dis}
\end{eqnarray}
Due to both $\psi^1$ and $\psi^2$ are monotone, we have  both $\Pr[\psi^1(v)(i) = j| v]$ and $\Pr[\psi^2(v)(i) = j| v]$ are non-increasing function of $v$ on the product space of vectors that satisfy $v(i)=j$. By the FKG inequality, we have
\begin{eqnarray}&&\mathbb{E}_{v\sim q}[\Pr[\psi^1(v)(i) = j] \Pr[\psi^2(v)(i) = j]| v(i)=j] \\
&&\geq \mathbb{E}_{v\sim q}[\Pr[\psi^1(v)(i) = j]| v(i)=j] \mathbb{E}_{v\sim q}[\Pr[\psi^2(v)(i) = j]| v(i)=j] \\
&& = \Pr[\psi^1(v)(i) = j| v(i)=j]\Pr[\psi^2(v)(i) = j| v(i)=j]\label{eq:disa}
\end{eqnarray}

(\ref{eq:dis}) and (\ref{eq:disa}) imply that
\begin{eqnarray}
\Pr[\psi(v)(i) = j| v(i)=j] &\geq& \Pr[\psi^1(v)(i) = j| v(i)=j]\Pr[\psi^2(v)(i) = j| v(i)=j]\\
&\geq& \alpha_1\alpha_2
\end{eqnarray}
The second inequality is due to the assumption that $\psi^1$ is an $\alpha_1$-CRS with regards to $q$ and $\psi^2$ is an $\alpha_2$-CRS with regards to $q$.

We next prove the monotonicity of $\psi$. For each $u, v \in [0;B]^I$
such that $u(i) = v(i)$ and $u \leq v$, we have
\begin{eqnarray}\Pr[\psi(u)(i) = u(i)] &=& \Pr[\psi^1(u)(i) = u(i)]\Pr[\psi^2(u)(i) = u(i)]\\
&\geq& \Pr[\psi^1(v)(i) = u(i)]\Pr[\psi^2(v)(i) = u(i)] \\
&=& \Pr[\psi(v)(i) = v(i)]
\end{eqnarray}
The inequality is due to the assumption that both $\psi^1$ and $\psi^2$ are monotone. $\Box$

\subsection{Algorithm Design}
\begin{algorithm}[hptb]
\caption{Inner and Outer Constrained Adaptive Policy $\pi^{io}$}
\label{alg:LPPX}
\begin{algorithmic}[1]
\STATE $A=\emptyset; i=1; j=1$.
\STATE compute a solution $y$ for \textbf{P1} by the
stochastic continuous greedy algorithm with stopping time $l=\min\{\beta, 1/4\} $ and step size $\delta=o(n^{-3})$
\FOR {$i\in I$} \label{ln:2}
\STATE add $i$ to $R^{io}$ with probability $\overline{y}(i)$
\ENDFOR
\STATE apply an outer constraint-specific monotone $(\beta, \gamma)$-balanced CRS  $\chi^{io}$ to $R^{io}$ to obtain a subset of items $\chi^{io}(R^{io})$ which satisfies the outer constraint \label{ln:3}
\FOR {$i\in \chi^{io}(R^{io})$}
\STATE sample a number $t$ from  $[C-c_i(B)]$ with probability $y(i,t)/\overline{y}(i)$
\STATE $t^{io}(i)\leftarrow t$
\ENDFOR
\STATE $\sigma^{io} \leftarrow$ sequence of items in $\chi^{io}(R^{io})$ sorted in a nondecreasing order of $t^{io}(i)$, breaking ties with the least index tie breaking rule
\FOR {$i\in I$}
\IF {$C' \leq t^{io}(\sigma^{io}_i)$} \label{ln:1}
\STATE select $\sigma^{io}_i$ and observe $\phi(\sigma^{io}_i)$
\STATE $C'=C' + c_{\sigma^{io}_i}(\phi(\sigma^{io}_i))$
\ENDIF
\ENDFOR
\end{algorithmic}
\end{algorithm}


Assume there exists  a monotone $(\beta, \gamma)$-balanced CRS for $\mathcal{I}^{out}$. Now we are ready the present the design of our \emph{Inner and Outer Constrained Adaptive Policy} $\pi^{io}$ (Algorithm \ref{alg:LPPX}). Our policy is composed of three steps, we next introduce each of those three steps in details.
\begin{enumerate}
\item Compute a solution $y$ for \textbf{P1} by the stochastic continuous greedy algorithm with stopping time $l=\min\{\beta, 1/4\} $ and step size $\delta=o(n^{-3})$.
\item Generate a random set $R^{io}$ by including each item $i$ with probability $ \overline{y}(i)$. Then apply a  monotone $(\beta, \gamma)$-balanced CRS $\chi^{io}$ to obtain a subset of items $\chi^{io}(R^{io})$ which satisfies the outer constraint.
\item Sample a number $t$ from  $[C-c_i(B)]$ with probability $y(i,t)/\overline{y}(i)$ for each $i\in \chi^{io}(R^{io})$. Let $\sigma^{io}$ denote a sequence of items in $\phi^{io}(R^{io})$ sorted in a nondecreasing order of $t^{io}(i)$.
\begin{enumerate}
\item Add $\sigma^{io}_1$ to the solution and observe $\phi(\sigma^{io}_1)$.
\item Starting with $i=2$. If $\sum_{i\in[i-1]} {c_{\sigma^{io}_i}(\phi(\sigma^{io}_i))} \leq t^{io}(\sigma^{io}_i)$, add $\sigma^{io}_i$ to the solution and observe $\phi(\sigma^{io}_i)$; otherwise, set $c_{\sigma^{io}_i}(\phi(\sigma^{io}_i))=0$. Repeat this step with the next item $i\leftarrow i+1$. This process continues  until all items from $\sigma^{io}$ have been visited.
\end{enumerate}
\end{enumerate}

\subsection{Performance Analysis}
This section is devoted to proving the approximation ratio of $\pi^{io}$. 
%
Recall that $y$ is obtained from solving problem \textbf{P1} using  the stochastic continuous greedy algorithm with stopping time $l=\min\{\beta, 1/4\} $ and step size $\delta=o(n^{-3})$. Then Lemma \ref{lem:kao} implies the following Corollary.
\begin{corollary}
\label{coro:kao}
Let $\pi^{opt}$ denote the optimal policy of our original problem. If we apply the stochastic continuous greedy algorithm
with stopping time $l=\min\{\beta, 1/4\} $ and step size $\delta=o(n^{-3})$ to solve \textbf{P1}, then the algorithm outputs a solution
 $y$ such that $\overline{y}\in \min\{\beta, 1/4\} \cdot P_{\mathcal{I}^{out}}$,   $\forall t\in [C]: \sum_{i\in I}\mathbb{E}[\min\{c_i(\phi(i)), t\}]\sum_{t'\in[t]}y(i,t')\leq  \min\{\beta, 1/4\}\cdot 2t$ and  $F(\overline{y})\geq (1-e^{-\min\{\beta, 1/4\}}-o(1))f_{avg}(\pi^{opt})$.
\end{corollary}

We next focus on proving that if there exists  a monotone $(\beta, \gamma)$-balanced CRS for $\mathcal{I}^{out}$, then $f_{avg}(\pi^{io})\geq (1-\min\{2\beta, 1/2\})\gamma F(\overline{y})$. This together with Corollary \ref{coro:kao} implies that $f_{avg}(\pi^{io})\geq (1-\min\{2\beta, 1/2\})\gamma (1-e^{-\min\{\beta, 1/4\}}-o(1))f_{avg}(\pi^{opt})$.

Consider a random vector $v\in [0; B]^I$ such that, for each $i\in I$, $v(i)$ is determined independently as $j \in [0; B]$ with probability
\begin{eqnarray}
\label{eq:ms}
h(i, j)= p_i(j)\cdot\overline{y}(i)
\end{eqnarray}
where we define $p_i(0)=\frac{1-\overline{y}(i)}{\overline{y}(i)}$ for each $i\in I$. Let $R(v)=\{i | i\in I \mbox{ and } v(i)\neq 0\}$. For the purpose of analyzing the performance  of Algorithm \ref{alg:LPPX}, we introduce three mapping functions: $\psi^a$, $\psi^b$, and $\psi^c$.

\begin{itemize}
\item  \textbf{Design of $\psi^a$.}    We  apply a monotone $(\beta, \gamma)$-balanced CRS $\chi^{io}$  used in Algorithm \ref{alg:LPPX} to $R(v)$ and obtain a set $\chi^{io}(R(v))\subseteq R(v)$.  We set $\psi^a(v)(i)=0$ for all $i\in I\setminus \chi^{io}(R(v))$, and set $\psi^a(v)(i)=v(i)$ for all $i\in \chi^{io}(R(v))$. 

\item \textbf{Design of  $\psi^b$.} Sample a \emph{starting time} $t(i)$ from  $[C-c_i(B)]$ with probability $y(i,t)/\overline{y}(i)$ for each $i\in R(v)$. Let $\sigma(v)$ denote the sequence of items in $R(v)$ sorted in a nondecreasing order of $t(i)$, breaking ties with the least index tie breaking rule. Let $\sigma(v)_{\leq t(i)}$ denote the sequence of items whose starting time is no later than $t(i)$.  
     For each  $i\in I\setminus R(v)$, we set $\psi^b(v)(i)=0$.  For each $i \in R(v)$, we set $\psi^b(v)(i)=v(i)$,  if
    \begin{eqnarray}\label{eq:aaa}
    \sum_{i'\in \sigma(v)_{\leq t(i)}\setminus \{i\}}^{i-1}c_{i'}(v(i'))\leq t(i)
    \end{eqnarray} and set $\psi^b(v)(i)=0$ otherwise.

\item \textbf{Design of $\psi^c$.} The third mapping function $\psi^c$ takes the intersection of   $\psi^a$ and $\psi^b$: First apply $\psi^a$ and $\psi^b$ to $v$ separately to obtain $\psi^a(v)$ and $\psi^b(v)$, then generate $\psi^c(v)$ as follows.
\begin{equation*}\mbox{For all } i\in I: \psi^c(v)(i)=
\begin{cases}
v(i)\quad &\mbox{if } \psi^a(v)(i)=v(i)\mbox{ and }\psi^b(v)(i)=v(i)\\
0 \quad &\mbox{otherwise }
\end{cases}
\end{equation*}
\end{itemize}

Before presenting the main theorem of this paper, we first provide several technical lemmas. The first four lemmas are used to lower bound the expected utility of $\psi^c(v)$ with regards to $h$. The fifth lemma shows that the expected utility of our policy is lower bounded by the expected utility of $\psi^c(v)$ with regards to $h$. Combing these two results, we are able to derive a lower bound on the expected utility of our policy.
\begin{lemma}
\label{lem:new1}
$\psi^a$ is a monotone $\gamma$-CRS with regards to $h$.
\end{lemma}
\emph{Proof:} We first prove that $\psi^a$ is a $\gamma$-CRS with regards to $h$. Recall that we set $\psi^a(v)(i)=0$ for all $i\in I\setminus S$, and set $\psi^a(v)(i)=v(i)$ for all $i\in \chi^{io}(R(v))$. It follows that for all $i\in I$ and all $j\in [B]$, we have
\begin{eqnarray}
\label{eq:2}
\Pr[\psi^a(v)(i) = j| v(i)=j] = \Pr[i\in \chi^{io}(R(v))| i\in R(v)]
 \end{eqnarray}
 Based on the definition of monotone $(\beta, \gamma)$-balanced CRS, we have $\Pr[i\in \chi^{io}(R(v))| i\in R(v)] \geq \gamma$. Hence, $\Pr[\psi^a(v)(i) = j| v(i)=j] \geq \gamma$. Next we prove that $\psi^a$ is monotone. Consider any two vectors $u, v \in [0; B]^I$ such that $u(i) = v(i)$ and $u \leq v$, we have $R(u) \subseteq R(v)$. Based on the definition of monotone $(\beta, \gamma)$-balanced CRS, we have $\Pr[i\in \chi^{io}(R(v))| i\in R(u)] \geq \Pr[i\in \chi^{io}(R(v))| i\in R(v)]$. Together with (\ref{eq:2}), we have $\Pr[\psi^a(u)(i) = j| v(i)=j] \geq \Pr[\psi^a(v)(i) = j| v(i)=j]$ for all $i\in I$ and  $j \in [B]$. $\Box$

 \begin{lemma}
 \label{lem:new2}
$\psi^b$ is a monotone $(1-\min\{2\beta, 1/2\})$-CRS with regards to $h$.
\end{lemma}
\emph{Proof:} The monotonicity of $\psi^b$ follows from Lemma 3 in \citep{fukunaga2019stochastic}. We next focus on proving that $\psi^b$ is a  $(1-\min\{2\beta, 1/2\})$-CRS. Consider any $i\in I$ and $k\in[C]$, we have
\begin{eqnarray}
&&\Pr[\psi^b(v)(i)=v(i)\mid  i\in R(v), t(i)=k] \\
&&= \Pr[\sum_{i'\in\sigma(v)_{\leq k}\setminus \{i\} }c_{i'}(v(i'))\leq k\mid i\in R(v), t(i)=k]\\
&&=\Pr[\sum_{i'\in\sigma(v)_{\leq k}\setminus\{i\}}\min\{c_{i'}(v(i')),k\}\leq k\mid i\in R(v), t(i)=k]
\end{eqnarray}  where
the probability considers two sources of randomness: one is the randomness in choosing $v$,
and the other source is the randomness in the generation of $t$. Because the event  that $\sum_{i'\in\sigma(v)_{\leq k}\setminus\{i\}}\min\{c_{i'}(v(i')),k\}\leq k$ is independent of the event that $i\in R(v)$ and $t(i)=k$, we have
\begin{eqnarray}
&&\Pr[\sum_{i'\in\sigma(v)_{\leq k}\setminus\{i\}}\min\{c_{i'}(v(i')),k\}\leq k\mid i\in R(v), t(i)=k] \\
&&= \Pr[\sum_{i'\in\sigma(v)_{\leq k}\setminus\{i\}}\min\{c_{i'}(v(i')),k\}\leq k]
\end{eqnarray}
We next provide a lower bound of $\Pr[\sum_{i'\in\sigma(v)_{\leq k}\setminus\{i\}}\min\{c_{i'}(v(i')),k\}\leq k]$. Observe that
\begin{eqnarray}
\mathbb{E}[\sum_{i'\in\sigma(v)_{\leq k}\setminus\{i\}}\min\{c_{i'}(v(i')),k\}]&&= \sum_{i'\in I\setminus\{i\}}\mathbb{E}[\min\{c_{i'}(v(i')),k\}]\Pr[t(i') \leq k]\\
&&= \sum_{i'\in I\setminus\{i\}}\mathbb{E}[\min\{c_{i'}(v(i')),k\}]\sum_{t\in[k]}y(i',t)
\\
 &&\leq \sum_{i'\in I}\mathbb{E}[\min\{c_{i'}(v(i')),k\}]\sum_{t\in[k]}y(i',t)\\
 &&\leq \min\{\beta, 1/4\}2k = \min\{2\beta, 1/2\}k
 \end{eqnarray}
The second inequality is due to $t(i)$ from  $[C-c_i(B)]$ with probability $y(i,t)/\overline{y}(i)$ for each $i\in R(v)$. The second inequality is due to   Corollary \ref{coro:kao} and the fact that $y$ is obtained from solving problem \textbf{P1} using  the stochastic continuous greedy algorithm with stopping time $l=\min\{\beta, 1/4\} $ and step size $\delta=o(n^{-3})$. Hence, $\Pr[\sum_{i'\in\sigma(v)_{\leq k}\setminus\{i\}}\min\{c_{i'}(v(i')),k\}> k] < \min\{2\beta, 1/2\}$ due to Markov inequality. It follows that $\Pr[\sum_{i'\in\sigma(v)_{\leq k}\setminus\{i\}}\min\{c_{i'}(v(i')),k\}\leq k]  =1- \Pr[\sum_{i'\in\sigma(v)_{\leq k}\setminus\{i\}}\min\{c_{i'}(v(i')),k\}> k]> 1- \min\{2\beta, 1/2\}$. $\Box$

The following lemma follows from Lemma \ref{lem:1}, Lemma \ref{lem:2} and Lemma \ref{lem:new4}.
 \begin{lemma}
 \label{lem:new3}
$\psi^c$ is a monotone $(1-\min\{2\beta, 1/2\})\gamma$-CRS with regards to $h$.
\end{lemma}

Theorem 4 in \cite{fukunaga2019stochastic} states that if $\psi^c$ is a monotone $\alpha$-CRS with respect to $h$,
then $\mathbb{E}_{v\sim h}[f(\psi^c(v))] \geq \alpha F(\overline{y})$. This together with Lemma \ref{lem:new3} implies the following lemma.

 \begin{lemma}
 \label{lem:1}
$\mathbb{E}_{v\sim h}[f(\psi^c(v))] \geq (1-\min\{2\beta, 1/2\})\gamma F(\overline{y})$.
\end{lemma}

We next show that the expected utility of $\pi^{io}$ is bounded by $\mathbb{E}_{v\sim h}[f(\psi^c(v))]$ from below.
\begin{lemma}
\label{lem:2}
 $f_{avg}(\pi^{io})\geq \mathbb{E}_{v\sim h}[f(\psi^c(v))]$.
\end{lemma}

\emph{Proof:} Recall that for any $S\subseteq I$ and any realization $\phi$, we use $\phi_S$ to denote a vector in $[0;B]^I$ such that $\phi_S(i)=\phi(i)$ if $i\in S$, and $\phi_S(i)=0$ otherwise. Let $\Phi_S$ denote a random realization of $S$.  As specified in Algorithm \ref{alg:LPPX}, $R^{io}$ is a random set that is obtained by including each item $i\in I$ independently with probability $\overline{y}(i)$. Thus, $\Phi_{R^{io}} \in [0; B]^I$ can be considered as a random vector such that, for each $i\in I$, $\Phi_{R^{io}}(i)$ is determined independently as $j \in [0; B]$ with probability
\begin{eqnarray}
\label{eq:ms1}
h(i, j)= p_i(j)\cdot\overline{y}(i)
\end{eqnarray}
where we define $p_i(0)=\frac{1-\overline{y}(i)}{\overline{y}(i)}$ for each $i\in I$. Note that the probability considers two sources of randomness: one is the randomness in choosing $R^{io}$, and the other source is the randomness of realization $\Phi$.

Now consider a fixed realization $v \in [0; B]^I$ of $\Phi_{R^{io}}$. Recall that in the design of $\psi^a$, we define $R(v)=\{i | i\in I \mbox{ and } v(i)\neq 0\}$. It is easy to verify that $R^{io}$ coincides with $R(v)$. For purpose of analysis, we further assume that $\chi^{io}(R^{io})$ coincides with $\chi^{io}(R(v))$. Moreover, for each $i\in \chi^{io}(R^{io})$,  we assume that $t(i)$ (the sampled starting time of $i$ in the implementation of $\psi^b$) coincides with $t^{io}(i)$ (the sampled starting time of $i$ in Algorithm \ref{alg:LPPX}).  This assumption indicates that $\sigma^{io}$ (a sorted sequence of items as specified in Algorithm \ref{alg:LPPX}) is a subsequence of $\sigma(v)$ (a sorted sequence of items as specified in the design of $\psi^b$) due to $\chi^{io}(R^{io}) \subseteq R^{io}$.

We next define a new mapping function $\psi^{io}$ as follows:
\begin{equation*}\mbox{For all } i\in I: \psi^{io}(v)(i)=
\begin{cases}
v(i)\quad &\mbox{if } i \mbox{ is selected by } \pi^{io} \mbox{ conditional on } v\\
0 \quad &\mbox{otherwise }
\end{cases}
\end{equation*}
Note that for $i\in I$ to be selected by $\pi^{io}$ conditional on  $v$, it must satisfy $i\in \chi^{io}(R^{io})$ as well as  the condition defined in Line \ref{ln:1} of Algorithm \ref{alg:LPPX} which can be written as
\begin{eqnarray}\label{cod}
\sum_{i'\in \sigma^{io}_{\leq t(i)}\setminus\{i\}: i'\mbox{ is selected by }\pi^{io}}c_{i'}(v(i'))\leq t(i)
\end{eqnarray} where  $\sigma^{io}_{\leq t(i)}$ denotes the subsequence of $\sigma^{io}$ by including all items whose starting time is no later than $t(i)$. It is easy to verify that $f_{avg}(\pi^{io}) = \mathbb{E}_{v\sim h}[f(\psi^{io}(v))]$. We next focus on proving that $\mathbb{E}_{v\sim h}[f(\psi^{io}(v))]\geq \mathbb{E}_{v\sim h}[f(\psi^c(v))]$.


According to the design of $\psi^c$, for each $i\in I$, $\psi^c(v)(i) = v(i)$ if and only if $i\in \chi^{io}(R(v))$ and condition (\ref{eq:aaa}) is satisfied. 
Given that $\sigma^{io}$ is a subsequence of $\sigma(v)$, condition (\ref{eq:aaa}) is stronger than the condition (\ref{cod}). This together with the fact that  $\chi^{io}(R^{io})=\chi^{io}(R(v))$ implies that for each $i\in I$, $\psi^c(v)(i) = v(i)$ implies that  $\psi^{io}(v)(i) = v(i)$.
 Hence, we have $\psi^{io}(v) \geq \psi^c(v)$, which implies that $f(\psi^{io}(v)) \geq f(\psi^c(v))$. It follows that $\mathbb{E}_{v\sim h}[f(\psi^{io}(v))] \geq \mathbb{E}_{v\sim h}[f(\psi^c(v))]$. This finishes the proof of this lemma due to $f_{avg}(\pi^{io}) = \mathbb{E}_{v\sim h}[f(\psi^{io}(v))]$. $\Box$

Corollary \ref{coro:kao}, Lemma \ref{lem:1}, and Lemma \ref{lem:2} imply the following main theorem.
 \begin{theorem}Assume there exists  a monotone $(\beta, \gamma)$-balanced CRS for $\mathcal{I}^{out}$, $f_{avg}(\pi^{io})\geq (1-\min\{2\beta, 1/2\})\gamma(1-e^{-\min\{\beta, 1/4\}}-o(1))f_{avg}(\pi^{opt})$.
 \end{theorem}

\subsection{Completing the Last Piece of the Puzzle: Discussion on $\beta$ and $\gamma$}
As the approximation ratio of $\pi^{io}$ is depending on the values of  $\beta$ and $\gamma$, we next discuss some practical outer constraints under which $\beta$ and $\gamma$ are well defined. 
In \cite{chekuri2014submodular}, they present monotone $(\beta, \gamma)$-balanced CRSs for a wide range of practical constraints including (multiple) matroid constraints, knapsack constraints, and their intersections. If  $\mathcal{I}^{out}$ is the intersection of a fixed number of knapsack constraints, there exists a $(1 -\epsilon, 1 - \epsilon)$-balanced CRS. If  $\mathcal{I}^{out}$ is induced by a matroid constraint, there exists a $(b, \frac{1-e^{-b}}{b})$-balanced CRS for any $b\in(0,1]$. We can use their results as subroutines in  Algorithm \ref{alg:LPPX} to handle a variety of outer constraints.
\bibliographystyle{ormsv080}
\bibliography{social-advertising-1}

\end{document}